\documentclass[10pt,twocolumn,letterpaper]{article}

\usepackage{cvpr}
\usepackage{times}
\usepackage{epsfig}
\usepackage{graphicx}
\usepackage{amsmath}
\usepackage{amssymb}

\usepackage[pagebackref=true,breaklinks=true,letterpaper=true,colorlinks,bookmarks=false]{hyperref}

 \cvprfinalcopy 

\def\cvprPaperID{****} 

\ifcvprfinal\fi
\begin{document}

\title{Classifying Cycling Hazards in Egocentric Data}

\author{Jayson Haebich\\
eXtended Reality Lab\\
City University Hong Kong\\
{\tt\small jhaebich@cityu.edu.hk}
\and
Christian Sandor\\
eXtended Reality Lab\\
City University Hong Kong\\
{\tt\small csandor@cityu.edu.hk}

\and
Alvaro Cassinelli\\
eXtended Reality Lab\\
City University Hong Kong\\
{\tt\small acassine@cityu.edu.hk}
}
\maketitle

\begin{abstract}

\begin{figure}
\includegraphics[width=0.95\linewidth]{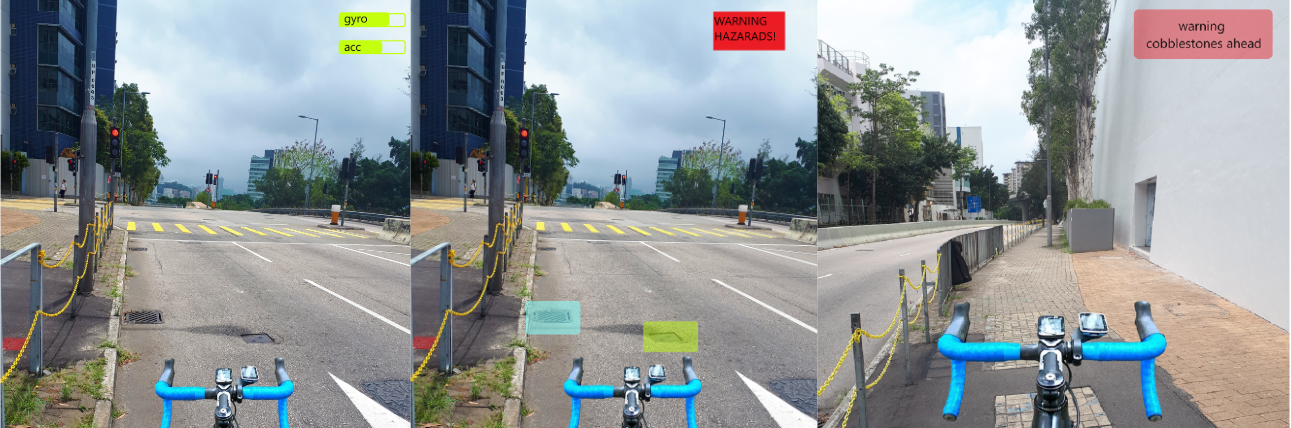}
\caption{using the proposed data set}
\end{figure}

   This proposal is for the creation and annotation of an egocentric video data set of hazardous cycling situations. The resulting data set will facilitate projects to improve the safety and experience of cyclists. Since cyclists are highly sensitive to road surface conditions and hazards they require more detail about road conditions when navigating their route. Features such as tram tracks, cobblestones, gratings, and utility access points can pose hazards or uncomfortable riding conditions for their journeys. Possible uses for the data set are identifying existing hazards in cycling infrastructure for municipal authorities, real time hazard and surface condition warnings for cyclists, and the identification of conditions that cause cyclists to make sudden changes in their immediate route. 

\end{abstract}
\section{Introduction}

This proposal is for the creation and annotation of an egocentric video data set of hazardous cycling situations. This data set will consist of egocentric video from a cyclist with associated data from an Inertial Measurement Unit (IMU) that indicates when the cyclist experienced sudden stopping or movement. The resulting data set will facilitate projects to improve safety and the experience of cyclists who are one of the most vulnerable groups travelling on roads \cite{department_of_transport_reported_2016}. Since cyclists are more sensitive to the surface conditions and hazards they require more detail about road conditions when navigating their route. Features such as tram tracks, cobblestones, gratings, and utility access points can pose hazards or uncomfortable riding conditions for their journeys. While there are data sets of egocentric videos for vehicles \cite{rateke_road_2019}, there is a lack of egocentric data sets for cyclists who navigate roads differently to drivers. Current public egocentric cycling data sets \cite{mahjourian_unsupervised_2018} only cover a limited range of cycling infrastructures and are missing many of the common hazards that cyclists face when they are cycling.

\section{Existing Data sets}
There are some data sets for egocentric vehicle video \cite{rateke_road_2019} but even though cyclists often share the road with cars they navigate roads differently. Existing vehicle based egocentric data sets are not appropriate since cyclists travel at a slower speed and in a different road position than those in a car. This means they have a different range of vision and are looking more closely at the immediate surface conditions compared with drivers. 

Existing egocentric cycling data sets only cover a limited range of cycling infrastructures typically found in the US \cite{mahjourian_unsupervised_2018}. This data set covers only roads, on road bike lanes and bike ways in a typical US city and does not include a wider range of cycling infrastructures found in other countries such as separated bike lanes, shared pedestrian and cycling paths. The data set also does not include a wide range of obstacles such as tram tracks, zebra/pedestrian crossings,cobblestone surfaces which are more commonly found in cities outside the US. 

Researchers have used KITTI, Carina and RTRK to classify road surfaces for vehicles into the following classifications; asphalt, asphalt(poor quality), paved and unpaved\cite{rateke_road_2019}. These classifications work for vehicles but are insufficient for cyclists who are more sensitive to the type of road surface they are travelling on. Other researchers have used their own classification process on egocentric cycling footage to classify cycling infrastructure \cite{rippel_using_2017}. But this study was focused on the type of infrastructure, not the surface of the road or hazards encountered. This data set was also not made public. 

\section{Data set Uses}
The resulting data set will be useful for those involved with the development and maintenance of cycling infrastructure. It will provide information on what kind of situations cause sudden changes or uncomfortable conditions for cyclists, enabling the better planning of infrastructure for cyclists. For example a city municipality can look at the resulting data and then at their own bike lanes to understand where dangerous conditions exist for cyclists or what are better types of surfaces to use when constructing bike lanes.

Another use is training machine learning models to identify hazards in video footage automatically. This can enable local cyclists or advocacy groups to find hazardous conditions within a city cycling network and push authorities to make changes to improve safety. An example is identifying what types of surfaces cause cyclists to brake suddenly or lose control more frequently, and where within a city certain hazards are more prevalent. 

Another use is pairing the visually identified hazards or conditions with the data from the IMU. This can enable a trained model to identify a more complex identification of current conditions without the use of a camera. If deployed on a larger scale such as with a bike sharing network then a complex analysis of the type of preferred paths and surfaces could be created using IMU data alone.

\section{Proposed Method}

The video and IMU data will be collected from cyclists via a customised helmet containing a raspberry pi with a webcam type camera and IMU that records to an SD card. The helmet will be worn by participants who will cycle through their daily commute or other common cycling routes in their city. The data will be collected across 3 countries (Hong Kong, Australia and the United Kingdom) with a total of 50 hours footage to be recorded. Several of these helmets will be made and sent to people to record data to increase the amount of data recording. The aim is to cover as many different types of cycling infrastructure as possible, including shared inner city roads, dedicated bike lanes, bike ways and leisure paths.

\begin{figure}
\includegraphics[width=0.85\linewidth]{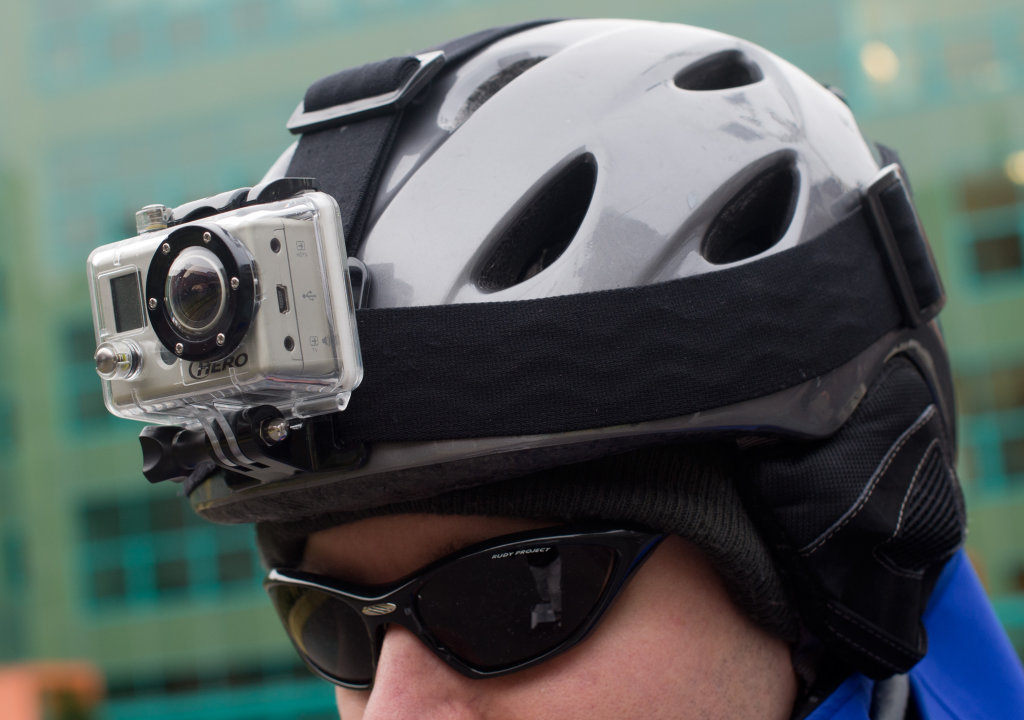}
\caption{example of proposed data collection helmet}
\end{figure}
Using the IMU information the video will be processed to find sections of 5-10 seconds where there was a sudden change in direction, sudden movement or some uncomfortable condition. Approximately 500 video sections will be extracted from the 50 hours of video. These sections will then be processed by MT to classify what hazards caused the sudden change in direction or movement. This is a more complex task that requires some analysis of the conditions by the MT who can deduce what cyclist to change their movement and is difficult for existing computer vision algorithms to handle.

\subsection{Privacy and legality}
 For each country where the egocentric cycling video is recorded it is legal to film on public roads without special permission \cite{city_of_melbourne_filming_2020} \cite{metropolitan_police_photography_2020} \cite{hong_kong_police_force_advisory_nodate}. To ensure privacy  personal information such as faces and license plates will be removed using software. Further processing of the data will create 5-10 second sections of video, that correspond to times when there was a sudden change of direction, sudden stopping, or heavy vibrations. The number of these video sections will vary but a total of around 10-15 per hour of footage will be the goal, if an insufficient amount is found then the threshold for identifying sudden changes of direction will be reduced to increase the number of video sections found. 

\subsection{Data storage}

Once collected and annotated the data will be stored on a server hosted by the XR Lab at City University of Hong Kong. This data will be publicly available for anyone to use or modify under the Creative Commons Attribution 4.0 International License. The video data will be stored as mov video files, IMU data will be saved as an XML file and the data classifications will be stored as XML. Permission has been granted from City University Hong Kong to release the annotated data as part of our research results and any participants who help to record the footage will be required to sign a release form.

\subsection{Timings}
This project will occur in two stages with the first stage being the data collection and the second stage being the data analysis with MT. The first data collection stage will take 3 months from when the project is confirmed and then the second data analysis stage taking another 3 months after this.

\subsection{Confirmation}
Permission has been granted from City University Hong Kong to release the annotated data as part of our research results.

\section{Amazon Mechanical Turk Usage}
\subsection{Task Overview}
The MT will be asked to watch a short videos between 5-10 seconds that  show a situation where there was a sudden change of movement or unpleasant cycling condition as indicated by the IMU. The analysis will be split into three tasks to reduce the amount of work per video analysis and receive more concise classifications from the users. Each of these tasks will be assigned to unique MTs.

\subsection{Budget requested}
 The total amount requested is \$2352 with the breakdown of this request as follows below. Each of the tasks will be repeated 5 times by a different MT to ensure the validity of the data.

\subsection{Task 1: Classify cause of movement or stopping }
The MT is allocated 2 minutes to complete the task as they may require to view the 5-10 second video section several times. The classifications for this task are:

\begin{table}[h]
\begin{tabular}{|l|c|l|}
\hline
 Tram tracks & Drains/Grating & cyclist  \\
 \hline
 No hazard&cobblestones  &  Pedestrian  \\
 \hline
  manhole & Vehicle &debris \\ 
 \hline
\end{tabular}
\caption{Task 1 Classifications}
\end{table}

\begin{table}[h]
\begin{tabular}{|l|l|}
\hline
Number of HITs & 500 \\
\hline
Reward per HIT (incl extra \%20) & \$0.384 \\
\hline
Assigned Workers per HIT & 5\\
\hline
Amazon Turk fee (20\%) & \$192\\
\hline
\hline
Total & \$1152\\
\hline
\end{tabular}
\caption{Task 1 breakdown}
\end{table}

\subsection{Task 2: Classify surface the user was travelling on during the sudden change of direction}

The MT is allocated 30 seconds to review the video and classify the road surface into one of the following classifications.
\begin{table}[h]
\begin{tabular}{|l|l|l|}
\hline
 Asphalt & Bad surface & Gravel  \\
 \hline
 Cobblestones & no change&    \\
 \hline
\end{tabular}
\caption{Task 2 Classifications}
\end{table}

\begin{table}[h]
\begin{tabular}{|l|c|}
\hline
Number of HITs & 500 \\
\hline
Reward per HIT & \$0.2 \\
\hline
Assigned Workers per HIT & 5\\
\hline
Amazon Turk fee (20\%) & \$100\\
\hline
\hline
Total & \$600\\
\hline
\end{tabular}
\caption{Task 2 breakdown}
\end{table}

\subsection{Task 3: Describe the hazard in the scene}
The MT will write a label or description for the video to describe what caused the sudden stopping. This will be used to validate the results received in the previous tasks.

\begin{table}[h]
\begin{center}
\begin{tabular}{|l|c|}
\hline
Number of HITs & 500 \\
\hline
Reward per HIT & \$0.2 \\
\hline
Assigned Workers per HIT & 5\\
\hline
Amazon Turk fee (20\%) & \$100\\
\hline
\hline
Total & \$600\\
\hline
\end{tabular}
\caption{Task 3 breakdown}
\end{center}
\end{table}

\subsection{Team Members}

\begin{itemize}
\item Jayson Haebich, Researcher eXtended Reality Lab
at City University of Hong Kong (corresponding person)
\item Christian Sandor, director of the eXtended Reality Lab
at City University of Hong Kong
\item Alvaro Cassinelli, director of the eXtended Reality
Lab at City University of Hong Kong
\end{itemize} 

\section{Conclusion}
The aim of this project is to create an egocentric data set of hazardous cycling conditions which will be useful for projects that facilitate cycling safety such as the creation and maintenance of cycling infrastructure. There is currently a lack of data sets relating to egocentric cycling conditions, and by using an IMU it is possible to collate a series of short videos that show  cycling hazards. The analysis of this data with MT will enable the classification of hazards that a cyclist encounters when they are undergoing some kind of sudden change of speed or direction.

{\small
\bibliographystyle{ieee_fullname}
\bibliography{EpicDataSetProposal}
}

\end{document}